\documentclass[journal]{IEEEtran}
\usepackage{cite}
\usepackage[pdftex]{graphicx}
\usepackage{amsmath}
\usepackage{algpseudocode}
\usepackage{algorithm}
\usepackage{subcaption}
\usepackage{tikz}

\usepackage{url}

\begin{document}
%
\title{Pulmonary Disease Classification Using Globally Correlated Maximum Likelihood:\\ an Auxiliary Attention mechanism for Convolutional Neural Networks}
%
%
%


\author{Edward Verenich,
        Tobias Martin,
        Alvaro Velasquez,
        Nazar Khan,
        and~Faraz Hussain
}

\maketitle

\begin{abstract}
Convolutional neural networks (CNN) are now being widely used for classifiying and detecting pulmonary abnormalities in chest radiographs. Two complementary generalization properties of CNNs, translation invariance and equivariance, are particularly useful in detecting manifested abnormalities associated with pulmonary disease, regardless of their spatial locations within the image. However, these properties also come with the loss of exact spatial information and global relative positions of abnormalities detected in local regions.  Global relative positions of such abnormalities may help distinguish similar conditions, such as COVID-19 and viral pneumonia. In such instances, a global attention mechanism is needed, which CNNs do not support in their traditional architectures that aim for generalization afforded by translation invariance and equivariance. Vision Transformers provide a global attention mechanism, but lack translation invariance and equivariance, requiring significantly more training data samples to match generalization of CNNs. To address the loss of spatial information and global relations between features, while preserving the inductive biases of CNNs, we present a novel technique that serves as an auxiliary attention mechanism to existing CNN architectures, in order to extract global correlations between salient features.
\end{abstract}

\begin{IEEEImpStatement}
We improve sensitivity of Convolutional Neural Networks (CNNs) using an auxiliary global attention mechanism (GCML) that enables CNNs to utilize global spatial information similar to Vision Transformers (ViTs). Our technique retains the benefits of spatial invariance and equivariance inherent to CNNs, while allowing spatial information of features to be used as discriminators.  GCML retains these inductive biases in data starved environments, which ViTs lack due their architecture, and hence require significantly more training data to achieve a similar level of generalization. Finally, we show impirically, that GCML improves the sensitivity of standard CNNs when classifying pulmonary conditions in chest X-rays. We provide all associated code, data, and models for reproducibility and improvement through further research.
\end{IEEEImpStatement}


%
\IEEEpeerreviewmaketitle

\section{Introduction}
%
%
%
%
\IEEEPARstart{W}{ith} the emergence of the COVID-19 pandemic, the use of Convolutional Neural Networks (CNNs) to detect presence of pulmonary diseases in medical imagery has quickly gained momentum, where a significant majority of approaches center around fine-tuning pre-trained CNNs on new data, as reported by Roberts et al. \cite{Roberts2021Nature}. The benefits of utilizing such techniques are intuitive, including the ability of CNNs to detect features that are difficult for humans to identify, learning certain correlations between positive cases, and the speed with which such predictions can be made in order to aid in timely diagnosis. CNNs have been shown to outperform individual radiologists in detecting pneumonia in chest X-rays as reported by Rajpurkar et al. \cite{Rajpurkar2017CheXNetRP}. In that work, X-ray based pneumonia diagnoses made by a group of four radiologists were compared to a custom CNN, using the F1 metric, where only one radiologist performed better than the model. \emph{This result prompted us to explore a possible mechasism of visual analysis of X-rays by a radiologist who outperformed the CNN, and whether that might translate into more accurate CNNs for classification of pulmonary diseases.}

Our hypothesis is that while human radiologists may not be as effective as CNNs at identifying individual salient features, their ability to quickly consider \emph{global spatial relations} between those features may play a factor. Our intution for this came from recent work by Borghesi et al. \cite{Borghesi_BRIXIA}, where an experimental scoring system for chest X-rays was used by radiologists to quantify and monitor disease progression in COVID-19 patients. The scoring worked by dividing the frontal X-ray of the lungs into six zones, where each zone was assessed with a quantitative score ranging from 0, representing no lung abnormalities, through 3, representing the highest severity of abnormalities. To obtain the final score, each region's scores were summed, and used as a measure of COVID-19 severity on the lungs. \emph{Our idea is to enable the CNN model to track abnormalities and their relations to each other across regions, and base our predictions on these learned spatial relationships and not just a cummulative score,} similar to what a human subject matter expert might do to distinguish different diseases.

In the remainder of this section, we briefly describe how image classification using CNNs works, including their strengths and limitations, with the goal to introduce the notion of \emph{additive} classification that we argue standard CNNs perform. We then discuss how certain positive generalization properties of CNNs limit their ability to account for global spatial correlations between regions of an image, thereby lacking the ability to utilize positional information as a discriminating factor.

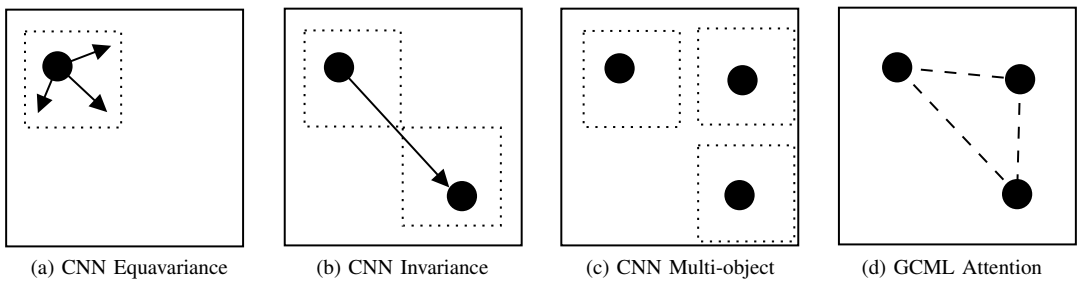
\begin{figure*}
  \centering
  \tikzset{every picture/.style={line width=0.75pt}} 

  \begin{tikzpicture}[x=0.75pt,y=0.75pt,yscale=-1,xscale=1]
  
  \draw   (21,81.5) -- (140.5,81.5) -- (140.5,201) -- (21,201) -- cycle ;
  \draw  [fill={rgb, 255:red, 0; green, 0; blue, 0 }  ,fill opacity=1 ] (39.75,110.13) .. controls (39.75,106.19) and (42.94,103) .. (46.88,103) .. controls (50.81,103) and (54,106.19) .. (54,110.13) .. controls (54,114.06) and (50.81,117.25) .. (46.88,117.25) .. controls (42.94,117.25) and (39.75,114.06) .. (39.75,110.13) -- cycle ;
  \draw  [dash pattern={on 0.84pt off 2.51pt}] (30.5,92.5) -- (79,92.5) -- (79,141) -- (30.5,141) -- cycle ;
  \draw    (46.88,110.13) -- (71.2,100.82) ;
  \draw [shift={(74,99.75)}, rotate = 519.0699999999999] [fill={rgb, 255:red, 0; green, 0; blue, 0 }  ][line width=0.08]  [draw opacity=0] (8.93,-4.29) -- (0,0) -- (8.93,4.29) -- cycle    ;
  \draw    (49,111.75) -- (69.81,131.2) ;
  \draw [shift={(72,133.25)}, rotate = 223.07] [fill={rgb, 255:red, 0; green, 0; blue, 0 }  ][line width=0.08]  [draw opacity=0] (8.93,-4.29) -- (0,0) -- (8.93,4.29) -- cycle    ;
  \draw    (46.88,110.13) -- (38.16,130.98) ;
  \draw [shift={(37,133.75)}, rotate = 292.68] [fill={rgb, 255:red, 0; green, 0; blue, 0 }  ][line width=0.08]  [draw opacity=0] (8.93,-4.29) -- (0,0) -- (8.93,4.29) -- cycle    ;
  \draw   (161.5,81) -- (281,81) -- (281,200.5) -- (161.5,200.5) -- cycle ;
  \draw   (441,80.5) -- (560.5,80.5) -- (560.5,200) -- (441,200) -- cycle ;
  \draw  [fill={rgb, 255:red, 0; green, 0; blue, 0 }  ,fill opacity=1 ] (181.75,110.63) .. controls (181.75,106.69) and (184.94,103.5) .. (188.88,103.5) .. controls (192.81,103.5) and (196,106.69) .. (196,110.63) .. controls (196,114.56) and (192.81,117.75) .. (188.88,117.75) .. controls (184.94,117.75) and (181.75,114.56) .. (181.75,110.63) -- cycle ;
  \draw  [dash pattern={on 0.84pt off 2.51pt}] (171.5,92) -- (220,92) -- (220,140.5) -- (171.5,140.5) -- cycle ;
  \draw  [fill={rgb, 255:red, 0; green, 0; blue, 0 }  ,fill opacity=1 ] (243.75,175.63) .. controls (243.75,171.69) and (246.94,168.5) .. (250.88,168.5) .. controls (254.81,168.5) and (258,171.69) .. (258,175.63) .. controls (258,179.56) and (254.81,182.75) .. (250.88,182.75) .. controls (246.94,182.75) and (243.75,179.56) .. (243.75,175.63) -- cycle ;
  \draw  [dash pattern={on 0.84pt off 2.51pt}] (221,141) -- (270.5,141) -- (270.5,190.5) -- (221,190.5) -- cycle ;
  \draw  [fill={rgb, 255:red, 0; green, 0; blue, 0 }  ,fill opacity=1 ] (463.25,110.63) .. controls (463.25,106.69) and (466.44,103.5) .. (470.38,103.5) .. controls (474.31,103.5) and (477.5,106.69) .. (477.5,110.63) .. controls (477.5,114.56) and (474.31,117.75) .. (470.38,117.75) .. controls (466.44,117.75) and (463.25,114.56) .. (463.25,110.63) -- cycle ;
  \draw  [fill={rgb, 255:red, 0; green, 0; blue, 0 }  ,fill opacity=1 ] (525.25,116.63) .. controls (525.25,112.69) and (528.44,109.5) .. (532.38,109.5) .. controls (536.31,109.5) and (539.5,112.69) .. (539.5,116.63) .. controls (539.5,120.56) and (536.31,123.75) .. (532.38,123.75) .. controls (528.44,123.75) and (525.25,120.56) .. (525.25,116.63) -- cycle ;
  \draw  [fill={rgb, 255:red, 0; green, 0; blue, 0 }  ,fill opacity=1 ] (523.75,174.63) .. controls (523.75,170.69) and (526.94,167.5) .. (530.88,167.5) .. controls (534.81,167.5) and (538,170.69) .. (538,174.63) .. controls (538,178.56) and (534.81,181.75) .. (530.88,181.75) .. controls (526.94,181.75) and (523.75,178.56) .. (523.75,174.63) -- cycle ;
  \draw  [dash pattern={on 4.5pt off 4.5pt}]  (470.38,110.63) -- (502.45,113.73) -- (532.38,116.63) ;
  \draw  [dash pattern={on 4.5pt off 4.5pt}]  (470.38,110.63) -- (530.88,174.63) ;
  \draw  [dash pattern={on 4.5pt off 4.5pt}]  (532.38,116.63) -- (530.88,174.63) ;
  \draw    (188.88,110.63) -- (242.47,169.04) ;
  \draw [shift={(244.5,171.25)}, rotate = 227.46] [fill={rgb, 255:red, 0; green, 0; blue, 0 }  ][line width=0.08]  [draw opacity=0] (8.93,-4.29) -- (0,0) -- (8.93,4.29) -- cycle    ;
  \draw   (301,81) -- (420.5,81) -- (420.5,200.5) -- (301,200.5) -- cycle ;
  \draw  [fill={rgb, 255:red, 0; green, 0; blue, 0 }  ,fill opacity=1 ] (323.25,111.13) .. controls (323.25,107.19) and (326.44,104) .. (330.38,104) .. controls (334.31,104) and (337.5,107.19) .. (337.5,111.13) .. controls (337.5,115.06) and (334.31,118.25) .. (330.38,118.25) .. controls (326.44,118.25) and (323.25,115.06) .. (323.25,111.13) -- cycle ;
  \draw  [fill={rgb, 255:red, 0; green, 0; blue, 0 }  ,fill opacity=1 ] (385.25,117.13) .. controls (385.25,113.19) and (388.44,110) .. (392.38,110) .. controls (396.31,110) and (399.5,113.19) .. (399.5,117.13) .. controls (399.5,121.06) and (396.31,124.25) .. (392.38,124.25) .. controls (388.44,124.25) and (385.25,121.06) .. (385.25,117.13) -- cycle ;
  \draw  [fill={rgb, 255:red, 0; green, 0; blue, 0 }  ,fill opacity=1 ] (383.75,175.13) .. controls (383.75,171.19) and (386.94,168) .. (390.88,168) .. controls (394.81,168) and (398,171.19) .. (398,175.13) .. controls (398,179.06) and (394.81,182.25) .. (390.88,182.25) .. controls (386.94,182.25) and (383.75,179.06) .. (383.75,175.13) -- cycle ;
  \draw  [dash pattern={on 0.84pt off 2.51pt}] (312.25,92.25) -- (360.75,92.25) -- (360.75,140.75) -- (312.25,140.75) -- cycle ;
  \draw  [dash pattern={on 0.84pt off 2.51pt}] (370,91) -- (418.5,91) -- (418.5,139.5) -- (370,139.5) -- cycle ;
  \draw  [dash pattern={on 0.84pt off 2.51pt}] (370,150) -- (418.5,150) -- (418.5,198.5) -- (370,198.5) -- cycle ;
  
  \draw (32,206) node [anchor=north west][inner sep=0.75pt]   [align=left] {{\footnotesize (a) CNN Equavariance}};
  \draw (176,206) node [anchor=north west][inner sep=0.75pt]   [align=left] {{\footnotesize (b) CNN Invariance}};
  \draw (451,206) node [anchor=north west][inner sep=0.75pt]   [align=left] {{\footnotesize (d) GCML Attention}};
  \draw (312,206) node [anchor=north west][inner sep=0.75pt]   [align=left] {{\footnotesize (c) CNN Multi-object}};

  \end{tikzpicture}
  
\caption{Examples of inductive biases inherent to Convolutional Neural Networks \emph{(a,b,c)}. Small translational shifts in a target object (black circle) within the receptive field of the kernel (dotted square) will shift the activation equally, and will still classify the target correctly due to \emph{(a)} translation equivariance. Major translational shift of the target object (black circle) to a new spatial position within the image will also result in the correct classification, regardless of its global position, due to \emph{(b)} translational invariance. Due to the scoring function, presence of multiple target objects \emph{(c)} within the image will result in a classification that the object is present in the image, regardless of quantity or spatial locations of the objects within the image. Our GCML attention mechansim \emph{(d)} uses spatial information of features, and their global interrelations, to distinguish between classes that can exhibit the same features in different locations, similar to Vision Transformers (ViTs) \cite{dosovitskiy2021an}, but with significantly fewer training samples needed by ViTs to achieve similar generalization. GCML achieves this by not performing tokenization of input images, but rather tokenizing class activation maps generated by the CNN with respect to the class of input images.}
\label{fig:inductive_biases_cnns}
\end{figure*}

\subsection{Image Classification with CNNs}
Yamashita et al. \cite{Yamashita_CNNs} provide an excellent overview of CNNs and their application in radiology, while we provide a general summary.  In a feed forward Convolutional Neural Network, image classification is performed by propogating input through a series of convolutional layers, where filters that were previously trained to recognize specific features of an image get activated and are used as input to subsequent layers. Typically, this process involves downsampling or reducing the mapping resolution of activated output with respect to input, as image data moves further through the layers. This is done by utilizing convolutional filters of size $k > 1$, but significantly less than the size of the input. By learning local correlation structures within the bounds of the window defined by $k$, particular convolutional layers learn specific local features. During inference, this allows the network to recognize objects or larger features, that are compositions of smaller features, which were recognized by earlier layers. After the last convolutional layer, 2 dimensional activation maps are flattened to 1 dimensional data and are passed to a fully connected layer that tallies the contributions of activated filters with respect to classes represented in the final connected layer. A classification can then be made by taking the maximum tally, or top K tallies for top K classification. Additional layers, such as a layer representing a \emph{SoftMax} function, can be utilized after the final connected layer to obtain class probabilities from the joint distribution of represented classes, but the core process described above is the same.

The use of convolutions in CNNs provides two important generalization properties with respect to image classification, \emph{translation equavarience and invariance}, where a convolution is equavarient to translation if an object in an image is spatially shifted, convolution's output is equally shifted. Invariance is a result of a position-independent pooling operation that follows a convolution, and results in a loss of absolute location, but enables the visual inductive prior of convolutional operators \cite{Osman_Translational_Eq}.  At a higher level, they enable features to be reliably detected regardless of their spatial location within an input image. 

CNNs became the go-to method for image classification after Krizhevsky et al. \cite{Krishevsky_2012} published their results on \emph{ImageNet} \cite{Deng_ImageNet}. Within the medical domain, CNNs started to be employed for diagnostic assistance of pulmonary diseases through classification of medical imagery, including CT scans and X-rays \cite{Wang_2017_CVPR} \cite{Sahiner_DLRAD}. With the rise of the COVID-19 pandemic, much work has been done in applying CNNs for image classification of medical imagery in order to rapidly detect pulmonary manifestations of COVID-19 in chest X-rays \cite{ALBAHRI20201381}. Much of this work, which we further discuss in Section \ref{sec:relatedwork}, utilizes CNNs for classifying chest X-rays in a manner we described above. For the purposes of this work, we refer to this method of image classification as \emph{standard} or \emph{additive}, because classification is performed by summing contributions of various filters at the final connected layer of the network.

\subsection{Limitations of Image Classification with CNNs}
Even with state-of-the-art results in image classification and object detection, CNNs have certain limitations relevant to image classification in certain domains.  For example, Hosseini et al. \cite{Hosseini_NegImages} report degraded image classification performance on images with reversed brightness, or \emph{negative} images. This may be mitigated by proper pre-processing of data, but it may signal that color channel information or texture may be learned by the network, along with shapes. Although not a limitation in itself, it introduces additional considerations when evaluating network performance on new and out-of-distribution data.

The main limitation that we address in this work has to do with the \emph{localized} perceptive fields of convolutional filters. Although CNNs exhibit translational invariance and equivariance, which improves generalization in many instances, it can also hurt generalization through loss of spatial information within the pooling layers of the CNN \cite{Sabour_NIPS2017}. The canonical example of this problem relates to classifying an image of a human face, where features such as nose, eyes, lips are identified, yet placing them in different parts of the image in a manner that does not resemble a face can still yield a face classification \cite{HJELMAS2001236}. Relating this to the chest radiographs domain, small manifestations of pulmonary disease are identified by a CNN, but their spatial interrelationships are mostly lost. In the natural imagery and radiology domains, the loss of spatial information, due to the contraints imposed by convolutional filter size and pooling, can result in incorrect classifications. Figure \ref{fig:inductive_biases_cnns} provides a visual intuition to the inductive biases that CNNs exhibit in terms of identifiying target objects or certain features, for example abnormalities in X-rays. While CNNs are particularly good at detecting abnormalities regardless of their spatial locations, the scoring function used for classification in standard CNNs, shown in Eq \ref{eq:class_score}, does not consider spatial locations nor spatial relations between features as discriminators.  We show in this work, that accounting for global spatial interrelation of features improves classification of pulmonary conditions.

\subsection{Contribution}
Our main goal in this work is to provide empirical evidence to the hypothesis that accounting for global correlations between activated regions of an image improves classification of pulmonary conditions in chest radiographs. This improvement also extends to image classification domains where discriminating a target class involves accounting for global spatial correlations between features.   Our secondary goal is to show that the classification approach, using our novel Globally Correlated Maximum Likelihood (GCML) auxiliary attention mechanism, is competitive to standard CNN classification, while utilizing significantly fewer model parameters, less computational resources, and much fewer training samples. To that end, our contributions are the following:

\begin{itemize}
 
  \item \emph{We developed a novel auxiliary attention mechanism, GCML, that is utilized with existing Convolutional Neural Network architectures in order to account for global spatial correlations between salient features of represented classes.}
  \item \emph{Using GCML, we show competitive results for image classification on a benchmark dataset, CIFAR-10, using significantly less model parameters and computation, while not utilizing any pre-training on samples in relevant domains.}
  \item \emph{We show that by utilizing the GCML attention mechanism, we improve image classification, specifically increasing model sensitivity or recall of pulmonary diseases in chest radiographs. This also includes effective generalization on a previously unseen dataset, suggesting that the GCML attention mechanism improves and complements visual inductive priors learned by CNNs by accounting for spatial relations.}
  \item \emph{Our results show that standard CNN and our GCML technique for image classification have particular strengths, and show potential utility when utilized as ensemble methods.}
  \item \emph{Finally, we provide an open source\footnote{\url{https://gitlab.com/verenich/gcmlpub}} reference implementation of our technique, allowing further research into its improvement and utility.}
\end{itemize}

The remainder of this paper is structured as follows.  Section \ref{sec:relatedwork} describes work related to image classification of pulmonary diseases using medical imagery. Section \ref{sec:approach} outlines our approach, while briefly discussing work relevant to attention mechanisms. We report results of our experiments in Section \ref{sec:experiments}, including a benchmark dataset of natural imagery and two separate datasets of chest radiographs. Finally we conclude by discussing our results, limitations of our approach, and future research directions.

\section{Related Work}
\label{sec:relatedwork}

In this section we describe work related to pulmonary disease classification using chest X-rays or CT scans. Some approaches mentioned also include forms of weakly supervised localization, which in contrast to object detection in imagery, does not require that training labels be accompanied with spatial coordinates of objects to be detected. These types of training labels are also referred to as \emph{image level} labels, as they only provide information on whether certain target classes are present in the image, not their spatial locations. To the best of our knowledge and at the time of this work, CNN attention mechanisms have not been used for the purpose of improving image classification of pulmonary diseases.

Rahaman et al \cite{Rahaman_COVID_TL_DL} employed transfer learning to fine-tune several CNN architectures to classify chest X-ray images into three classes: COVID-19, Healthy, and Pneumonia. A total of 860 images were used in their study, which reported the VGG19 \cite{Liu_VGG} architecture having the best performance achieving an accuracy of 89.3 percent, average precision of 0.90, a recall of 0.89, and F1 score of 0.90. Khan et al. \cite{Khan_CoroNet} proposed CoroNet, a network based on the Xception \cite{Chollet_2017_CVPR} architecture for diagnosis of COVID-19 from chest X-rays. They used a dataset of X-ray images \cite{cohen2020covid} to train their network, achieving 95 percent accuracy when classifying images for COVID-19, Normal, and Pneumonia. Tamal et al. \cite{TAMAL2021115152} used their radiology classification model for COVID-19 classification with low severity, as scored by radiologists, on new data from patients at a local hospital, showing generalization to out-of-distribution data and achieving an overall accuracy of 90 percent. Kim et al. \cite{Kim_datasetComposition} reported relevant findings on the effect of dataset composition practices on classification performance.  The authors reported that higher classification performance was observed on datasets where data composing each class came from different sources. Heidari et al. \cite{Heidari_ImagePreprocess} showed that their image preprocessing scheme improved pulmonary disease classification in X-ray images. Similar diagnostic aid approaches to classifying COVID-19 were reported in \cite{Khuzani2020_COVID19}, \cite{Alruwaili_COVID19}, and \cite{Purohit2020_COVID19}, where the use of transer learning using a pre-trained Convolutional Neural Network architecture to perform multi-class classification was the common factor.

In addition to classifying images as representing pulmonary conditions, work has been proposed to provide some explainability of those classifications.  Tsiknakis et al. \cite{Tsiknakis_intrCOVID19} utilized the Inception \cite{Szegedy_InceptionV3} architecture to first classify X-ray images into specific diseases, and then applied a weakly supervised localization technique GradCAM \cite{selvaraju2017GradCAM} to identify regions within the image responsible for a particular classification. Wang et al. \cite{Wang_Covid_localization} proposed a similar approach to diagnose and localize disease manifistation in X-rays. Verenich et al. \cite{verenich2020improvingICMLA} proposed a method to reduce aleatoric uncertainty in weakly supervised localization that can arise from significant class overlap between features associated with similar pulmonary diseases. Gupta et al. \cite{Gupta_InstaCovNet} proposed an approach that classifies and performs weakly supervised localization using standard Class Activation Maps \cite{zhou2016learning}.

Recently, Transformers, originally proposed by Vaswani et al. \cite{NIPS2017_3f5ee243}, have shown state-of-the-art performance on natural language processing tasks. These ideas have been utilized in Vision Transformers (ViT) \cite{dosovitskiy2021an} to introduce positional attention mechanisms for image classification, attaining results comparable to state-of-the-art CNNs. However, ViTs lack translational invariance and equavarience of CNNs, thus require significanly more training data to generalize \cite{dosovitskiy2021an}. To the best of our knowledge, and at the time of this work, ViTs have not been used for classification of pulmonary diseases using X-rays. One possible reason for this is insufficient amount of training data, to achieve same levels of generalization as CNNs with the data that is currently available.


\section{Approach: GCML Attention Mechanism}
\label{sec:approach}
The goal of our attention mechanism is to preserve spatial interrelationships of activated regions in a given image $I$ relative to target classes $C$ learned by a convolutional neural network $G$. Our hypothesis is that localized pulmonary abnormalities, detected by the convolutional neural network in different regions, can yield additional discriminative power when their global interrelationships are considered. The main intuition for this is that translation invariance and equivariance properties of the CNNs result in class discrimination that is additive, with respect to activation strengths of convolutional filters, but not their spatial relation to each other.  This section describes the architecture of our approach, that accounts for spatial interrelations of salient features.

\subsection{Attention Type}
Our technique explores complementing convolutional neural networks with attention mechanisms. The main difference between prior work \cite{GCNet_Cao,DisentangeledNN_Yin,NonLocalNN_Wang,Ramachandran2019StandAloneSI} and GCML is that \emph{we do not alter the architecture of the CNN} using self-attention layers, but instead provide a separate auxiliary structure that is created using a pre-trained network.  In addition, the stochastic structure of GCML does not require it to be trained simultaneously with the CNN using \emph{backpropagation} as in \cite{Ramachandran2019StandAloneSI}, thus it does not have to be differentiable and is significantly faster to train. Finally, tokenization of input images into patches is also not required, instead we use downsampled output of the last convolutional layer, scaled by class weights from the final connected layer, to generate class activation maps, as proposed by Zhou et al. \cite{zhou2016learning}. These class activation maps are used as input to our GCML attention mechanism. In addition to being effective at performing weakly supervised localization, class activation maps were also used to distinguish overlapping regions of images that may belong to different but overlapping classes \cite{verenich2020improvingICMLA}. 

The work that is in closest alignment with our approach is the UL-Hopfield model \cite{hopfieldNetwork}, which uses pre-trained CNNs with an associative memory bank to perform image classification. Our approach differs in several ways: the UL-Hopfield auxiliary memory learns class-specific \textit{core patterns} from a pre-trained CNN in an unsupervised manner, while we use the CNN to generate a pattern using training labels. Second, the type of input to the attention function that is used to train their memory structure is extracted from the last pooling layer without weighting activated feature maps by a specific class, which is done in our approach during training. Finally, for experiments we use a CNN with 42x less parameters than the CNN they use for feature extraction, as we discuss in Section \ref{sec:experiments}.

\subsection{Input Features}

Formally, to compute the attention function input tensor $M_c$, where $c$ is a class represented in the CNN, we do the following: given an input image, let $f_k(x,y)$ be the activation of filter $k$ at the last convolutional layer of of a CNN $G$ and $(x,y)$ be the spatial location. During classification, for each filter $k$, a pooling layer outputs a global average $F_k$ defined as $\sum_{x,y} f_k(x,y)$, which is then used as input to the fully connected layer, giving us the class score $S_c$ in Eq \ref{eq:class_score}, where $w^c_k$ is a scalar weight indicating the importance of $F_k$ to class $c$.

\begin{equation}
  S_c = \sum_k w^c_k F_k
  \label{eq:class_score}
\end{equation}

\noindent To compute each spatial element of the class activation map \cite{zhou2016learning}, or 2-dimensional tensor $M_c$, we use

\begin{equation}
  M_c(x,y) = \sum_k w^c_k f_k(x,y)
  \label{eq:cam}
\end{equation}
\noindent The resulting tensor $M_c$ effectively splits the input image $I$ into regions or tokens, where entries $M^c_{i,j}$ represent activation intensities of those regions for class $c$.

Our technique would be equally applicable to be used with class activation maps generated by another weakly supervised localization technique called Grad-CAM, proposed by Selvaraju et al. \cite{selvaraju2017GradCAM}. It requires that we compute the gradients of output of the network with respect to feature map activations for each class $c$, making it slower than the standard CAM method for the purposes of training the GCML structure on large datasets.

\subsection{Attention Function}
\begin{figure}[]
  \centering
  \includegraphics[width=2.5in]{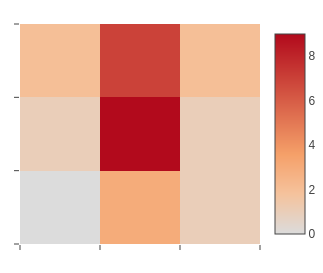}
  \caption{Example of a 2-dimensional tensor $M$, which is the same as the class activation map, that is used as an input to the attention function $Q$. The dimensions of $M$ are dictated by the mapping resolution of the convolutional layer that we use to compute the class activation map. Values at $M_{i,j}$ represent activation intensities for class $C$ after an image $I$ is passed through the network $G$.}
  \label{fig:cam_input}
\end{figure}

\begin{figure}[]
  \centering
  \includegraphics[width=2.5in]{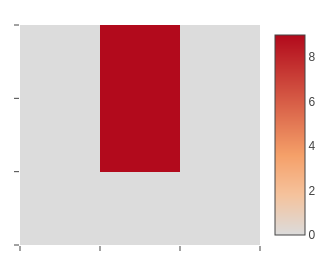}
  \caption{Example of a 2-dimensional tensor $M$ in its intermediate state after normalization and application of threshold $\tau$ as computed within attention function $Q$. Hyperparameter $\tau$ is also optimized during the training stage of the GCML structure and its optimized value is persisted for inference.}
  \label{fig:cam_tau}
\end{figure}

\begin{figure}[]
  \centering
  \includegraphics[width=3in]{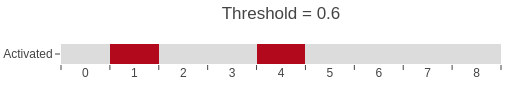}
  \caption{Intermediate state of tensor $M$ is flattened to a vector $B$ where entries $B_i$ represent bits that are set to 0 or 1. Converting this vector of bits to an integer yields our datastore key $K$, which would be 144 in the vector displayed. This value is the output of the attention function $Q$. Note that given the length $L$ of vector $B$, the number of possible entries $K_i$ is  $2^L $. Also, either big or little endianness can be used for bit arrangment, as long as it is consistent throughout training and inference.}
  \label{fig:cam_flat}
\end{figure}

Similar to the original work on transformers \cite{NIPS2017_3f5ee243}, our attention function is a mapping of a query $Q$, based on a datastore key $K$ to a retrieved value $V$. In our case, the query function $Q$ takes as input a real tensor $M$ of size $H \times W$, which maps it to a datastore key $K$, to retrieve likelihood value $V$. In training mode, when learning the GCML structure, key $K$ is used to update the value at that index, while during inference mode, it is used to retrieve $V$ at position $K$. In both cases, attention function $Q$ remains the same. The states of input, intermediate states, and output of $Q$ are illustrated in Figures \ref{fig:cam_input}, \ref{fig:cam_tau}, and \ref{fig:cam_flat}. Parameter $\tau$ is used as a threshold to convert input $M_c$ to its intermediate state shown in Fig \ref{fig:cam_tau} using Equation (\ref{eq:tau}).

\begin{equation}
  f_{\tau}(M_{i,j}) = 
  \begin{cases}
    1, & \text{if } M_{i,j} \geq \tau \\
    0, & \text{otherwise}
  \end{cases}
  \label{eq:tau}
\end{equation}

\noindent The full algorithm for obtaining datastore key $K$ from tensor $M_c$ using the attention function $Q$ is described in Algorithm \ref{alg:q}.

\begin{algorithm}
  \caption{Attention function Q}
  \begin{algorithmic}[1]
    \Procedure{Q}{$M_c$}\Comment{Computes $K$ from $M_c$}
      \State $M'_c \gets normalize(M_c)$
      \State $M''_c \gets f_{\tau}(M'_c)$ \Comment{Threshold $\tau$}
      \State $B \gets flatten(M''_c)$ \Comment{From 2-d to 1-d}
      \State $K \gets binToInt(B)$
      \State \textbf{return} $K$ \Comment{Datastore key $K$}

    \EndProcedure
  \end{algorithmic}
  \label{alg:q}
\end{algorithm}

\subsection{Training GCML}
The GCML datastore $S$ is a tensor with dimensions ${C \times P}$, where $C$ is the number of classes represented in network $G$ and $P$ is of size $2^L$, where $L$ is the length of the flattened vector $B$ from which key $K$ is derived. For a given class $y$, and possible activations $x$ of $M^y_{i,j}$, being in on or off states, \emph{each row} of $S$ represents a discrete \emph{Conditional Probability Distribution} of class $y$ given activations $x$, as shown in Equation (\ref{eq:jcpd}).

\begin{equation}
  S[y] = P(y | x_0, y | x_1, \dots, y | (x_0,x_1,\dots,x_L))
  \label{eq:jcpd}
\end{equation}

\noindent To compute these likelihood distributions for all classes, we utilize a fully trained convolutional neural network $G$ along with the training data set $D_T$, which was used to train $G$.

\begin{algorithm}
  \caption{GCML training procedure (1 epoch)}
  \begin{algorithmic}[1]
    \Procedure{update}{$S,D_T,G$}\Comment{Update likelihoods in S}
      \For{$i,l_c$ in $D_T$} \Comment{Image $i$, with label $l_c$}
        \State $M_c \gets G_m(i,l_c)$ \Comment{Compute $M_c$ via $G$ and Eq \ref{eq:cam}}
        \State $K \gets Q(M_c)$ \Comment{Compute key $K$}
        \State $S[l_c][K] \gets +1$ \Comment{Increment state}
      \EndFor
      \State $S \gets normalize(S)$ \Comment{Optional normalization}
      \State \textbf{return} $S$ \Comment{GCML store $S$}
    \EndProcedure
  \end{algorithmic}
  \label{alg:train}
\end{algorithm}

Algorithm \ref{alg:train} shows the training procedure for the GCML datastore. For simplicity, we show the procedure for single images sequentially. In practice however, we implement these procedures on batches of images provided by a dataloader. We also note that none of the weights in network $G$ are being updated during this procedure, and the network is set to evaluation mode. The procedure $G_m(i,l_c)$ involves propagating the image through network $G$, and computing $M_c$ for that image using class label $l_c$. Finally, the normalization procedure of $S$ is marked as optional because this normalization can also happen before inference using GCML is performed. One reason to hold off on normalization, is to allow $S$ to be further trained on additional data, as we will discuss later in this paper.

Another consdideration when training the GCML structure is to account for data transformations that were performed while training the original network $G$. For example random crops or flips along a horizontal or vertical axes may alter the class activation map, thus using more epochs with random transformations should improve the GCML structure in some domains, such as natural imagery of CIFAR-10.  We note however, that in a domain such as chest radiology, major transformations such as flips will actually degrade performance as orientation of X-ray images is consistent and generalizing to such tranformations is not needed and is actually harmful. Therefore, in order to improve generalization of attention mechanisms, appropriate transformations should be considered based on the target domains.

\subsection{Inference with GCML}
To perform inference using the GCML attention mechanism we utilize a trained Convolutional Neural Network $G$ along with the GCML datastore $S$. As mentioned earlier, the attention function $Q$ remains the same during training and inference, the main difference is that during inference we compute $M_c$ for \emph{all} classes represented in $G$ instead of just the class label provided with training data. Additionally, before inference is performed, we must make sure that the datastore $S$ is normalized, as the normalization step during training shown on line 7 in Algorithm \ref{alg:train} is optional, in order to enable further training.

\begin{algorithm}
  \caption{GCML inference procedure}
  \begin{algorithmic}[1]
    \Procedure{predict}{$I,G,S$}\Comment{Predict class on image $I$}
      \State $\vec{M_c} \gets G(I)$ \Comment{Tensors $M_c$ for all classes in $G$}
      \State $\vec{K_c} \gets Q(\vec{M_c})$ \Comment{Compute $K$ for all classes in $G$}
      \State $\vec{V_c} \gets lookup(S,\vec{K_c})$ \Comment{Get class likelihoods}
      \State $C_P \gets argmax(\vec{V_c})$ \Comment{Get Max Likelihood class}
      \State \textbf{return} $C_P$ \Comment{Return predicted class}
    \EndProcedure
  \end{algorithmic}
  \label{alg:infer}
\end{algorithm}

Algorithm \ref{alg:infer} shows the inference procedure of our approach, which expects datastore $S$ to be normalized.  For each image $I$, we propagate it through the convolutional neural network $G$, where we compute inputs $\vec{M_c}$ to the attention function $Q$, where each item in $\vec{M_c}$ is a class activation map computed for each class represented in $G$, given input $I$. In other words, single input $I$ will generate $N$ inputs to the attention function, where $N$ is the number of classes in $G$. The next step is to compute vector $\vec{K_c}$, where each entry $K_i$ is a key to a likelihood value representing class $i$. Class likelihood values $\vec{V_c}$ are then retrieved from $S$ and maximum value is taken to determine the class that input $I$ belongs to.

\begin{figure*}[!t]
  \centering
  \includegraphics[width=6.5in]{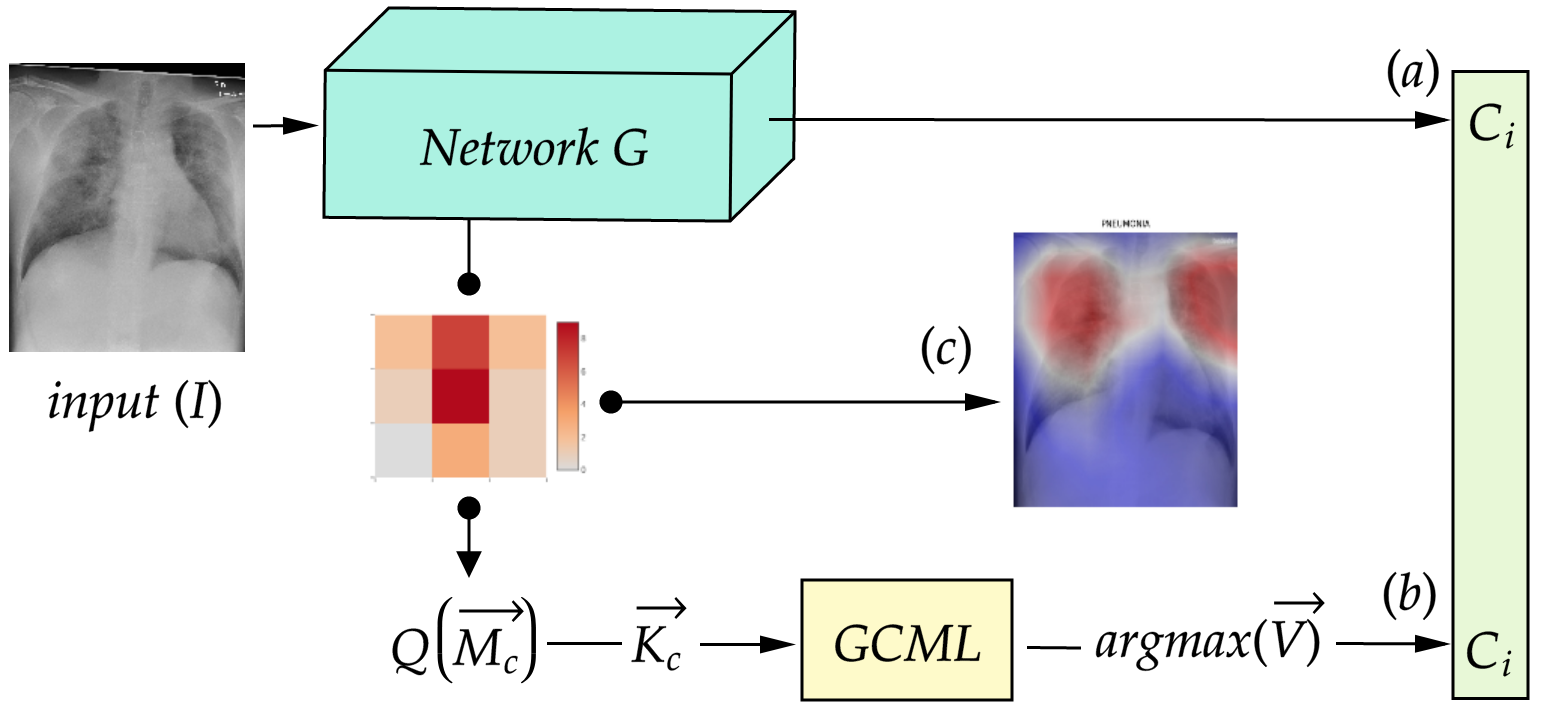}
  \caption{Inference flow using a Convolutional Neural Network $G$ and the auxiliary attention mechanism GCML. For a given input image $(I)$, outputs $(a)$ and $(b)$ represent classifications made by the Convolutional Neural Network through its standard classification layer $(a)$, and using the GCML attention mechanism $(b)$. Both classifications can be utlized as an ensemble in a single forward pass during inference. In addition, input tensor $M_c$, to the attention function $Q$, can be upsampled using bilinear sampling to obtain a heatmap that performs weakly supervised localization of relevant regions for a given class as shown with output $(c)$. We note that we include path $(c)$ as an example of weakly supervised localization that can be performed using $M_c$, which we do not perform in our experiments. The input to the attention function $Q$ is marked as a vector because 2-d tensors $M_c$ are computed for every class represented in network $G$, hence a vector of 2-d tensors is passed to $Q$ to compute a vector of class keys $\vec{K_c}$ that is then used to retrieve a vector of class likelihood probabilities $\vec{V}$ as shown in Algorithm \ref{alg:infer}. We also note that the block labeled as $GCML$ represents GCML datastore tensor $S$ as shown in Algorithm \ref{alg:train}.}
  \label{fig:gcml_flow}
\end{figure*}

Figure \ref{fig:gcml_flow} shows a full view of the inference procedure using the GCML attention mechanism, as well as the standard inference process using the convolutional neural network. As shown in the diagram, in addition to two classifications, labeled $(a)$ for standard CNN classification and $(b)$ for classification using the attention mechanism, $Q$ function input tensors $\vec{M_c}$ can also be used to perform weakly supervised localization $(c)$ by upsampling them to the same size as the input image to produce a heatmap of relevant regions for a given class.

\subsection{On Attention Input Size}
The dimensions of the input $M_c$ to the attention function $Q$ are determined by the final mapping resolution of the last convolutional layer of network $G$. For example, ResNet50, a version of a widely used convolutional architecture utilizing residual layers \cite{resnet_He}, has a final mapping resolution of $7 \times 7$, when used with input images that are $224 \times 224$. This would result in GCML datastores with $2^{49}$ entries for each class, as this is the number of possible binary combinations of threshold activated regions within $M_c$ of that size. Instead, we downsample the final resolution layer to a more manageable size, $4 \times 4$ and $5 \times 5$ as we will discuss in Section \ref{sec:experiments}.

\section{Experiments}\label{sec:experiments}
In this section we report the results of using our GCML attention mechanism to perform image classification. We performed experiments on three datasets: (1) dataset of natural imagery CIFAR10 \cite{cifar10}, (2) COVID-19 radiology dataset \cite{covidDataKaggle1} containing X-ray imagery of patients diagnosed with COVID-19, viral pneumonia, and no findings, (3) dataset of COVID-19, pneumonia, and no findings images taken from \cite{cohen2020covid}. The purpose of the third dataset is to assess generalization of our method to new data.

The CIFAR10 dataset contains 60000 $32 \times 32$ images representing 10 mutually exclusive classes, meaning each image belongs to only one class, which are: airplane, automobile, bird, cat, deer, dog, frog, horse, ship, and truck. Dataset authors note that automobile and truck classes do not overlap. The main reason we use CIFAR10 for our experiments is that the authors of the approach most similar to ours \cite{hopfieldNetwork} provide extensive evaluation results, as well as being the largest dataset evaluated in that work.

As the main data for our experiments, the COVID-19 radiology dataset \cite{covidDataKaggle1} contains a total of 2905 images, where COVID-19 cases are only represented in 219 of them, with the rest evenly distributed between viral pneumonia and images with no findings. This data set is particularly interesting to us as it represents both class imbalance and a data starved environment, as in such cases vision transformers do not outperform state-of-the-art CNN models \cite{dosovitskiy2021an} without pre-training on very large datasets in the similar domain.

To remain within reasonably accessible hardware constraints, all of our experiments were performed on a single machine with the following hardware characteristics: Intel Core i7 CPU, 64 GiB of RAM, NVIDIA RTX 2080 GPU. We used Ubuntu 20 as the operating system, NVIDIA CUDA 11 GPU acceleration library, and Pytorch 1.9.0 as our model implementation framework.

\subsection{CIFAR-10 Results}
Here we describe our experimental settings and results on the CIFAR-10 benchmark dataset. As mentioned earlier, the model we use as our feature extractor is significantly smaller than that used in \cite{hopfieldNetwork}, where a pretrained ResNet50 architecture was used. We implemented a smaller version of a residual network architecture, which contains only 787,482 parameters compared to 33,554,432 for ResNet50. Additionally, we did not perform resizing of input images to $224 \times 224$, but instead used the original input size of $32 \times 32$. The mapping resolution of the last convolutional layer in our network is $4 \times 4$ compared to $7 \times 7$ utilized in \cite{hopfieldNetwork}. The authors \cite{hopfieldNetwork} did not fine-tune their feature extractor on the CIFAR-10 dataset, while we trained ours from scratch for 80 epochs on a single GPU (Nvidia RTX 2080) taking around 15 minutes or about 0.25 GPU hours. The only data augmentation we performed was padding input images by 4 pixels and randomly cropping at the original size of $32 \times 32$. The training phase of the GCML auxiliary attention mechanism took 28 minutes on the same machine with several $\tau$ values. Training the $GCML$ structure with different values of $\tau$ does not require any retraining of the CNN feature extractor. The test partition of the CIFAR-10 dataset (10,000 images) was not used for either of training phases.  We also did not utilize a validation set during training of our feature extractor in order to replicate the training environment of \cite{hopfieldNetwork} as much as possible.

\begin{table}[!]
\renewcommand{\arraystretch}{1.3}
\caption{CIFAR-10 Results for image classification using our GCML approach, UL-Hopfield network \cite{hopfieldNetwork}, and state-of-the-art vision transformer model \cite{dosovitskiy2021an}. We include vision transformer models to illustrate efficiency of our approach in terms of number of parameters and computational cost. The two transformer models were pre-trained on 300M images using cloud TPUv3 with 8 cores for 8 days \cite{dosovitskiy2021an}, while our method took less than 1 hour on a single GPU. The numbers of trainable parameters are reported for each method according to standard practice.}

\centering
\begin{tabular}{|c|c|c|c|}
\hline
Model & \% Correct & Parameters & Extra Training Data\\
\hline
ViT-H/14 \cite{dosovitskiy2021an} & 99.5 & 632M & 300M\\
ViT-L/16 \cite{dosovitskiy2021an} & 99.42 & 307M & 300M\\
\hline
UL-Hopfield \cite{hopfieldNetwork} & 83.1 & 33.6M & 1.28M\\
\hline
Res4Cif GCML* & \textbf{85.5} & 0.79M & 0\\
\hline
\end{tabular}
\label{tab:cifar_results}
\end{table}

Table \ref{tab:cifar_results} shows our results marked with (*) along with reported results in \cite{hopfieldNetwork} and the state-of-the-art performance on the CIFAR-10 dataset reported in \cite{dosovitskiy2021an}. Only predictions made using the GCML data structure, marked $(b)$ in Figure \ref{fig:gcml_flow}, were used. Our methods outperforms \cite{hopfieldNetwork}, but we note that the training of their memory module was unsupervised, even though their pre-trained feature extractor was significantly larger.  One observation in these results is the significant reduction in parameters and extra training costs shown by our method, where our model has 800x fewer parameters than ViT-H/14 and 42x fewer parameters than UL-Hopfield while requiring no extra data to pre-train. Both ViT-H/14 and ViT-L/16 are vision tranformer models, which are pretrained on large datasets. ViT-L/16 model, is pre-trained on the ImageNet-21k dataset, and is trained on cloud TPUv3 with 8 cores for approximately 8 days \cite{dosovitskiy2021an}. Our method had a combined training time of less than 1 hour on a single machine with 1 GPU.

\begin{figure}[!]
  \centering
  \includegraphics[width=3.2in]{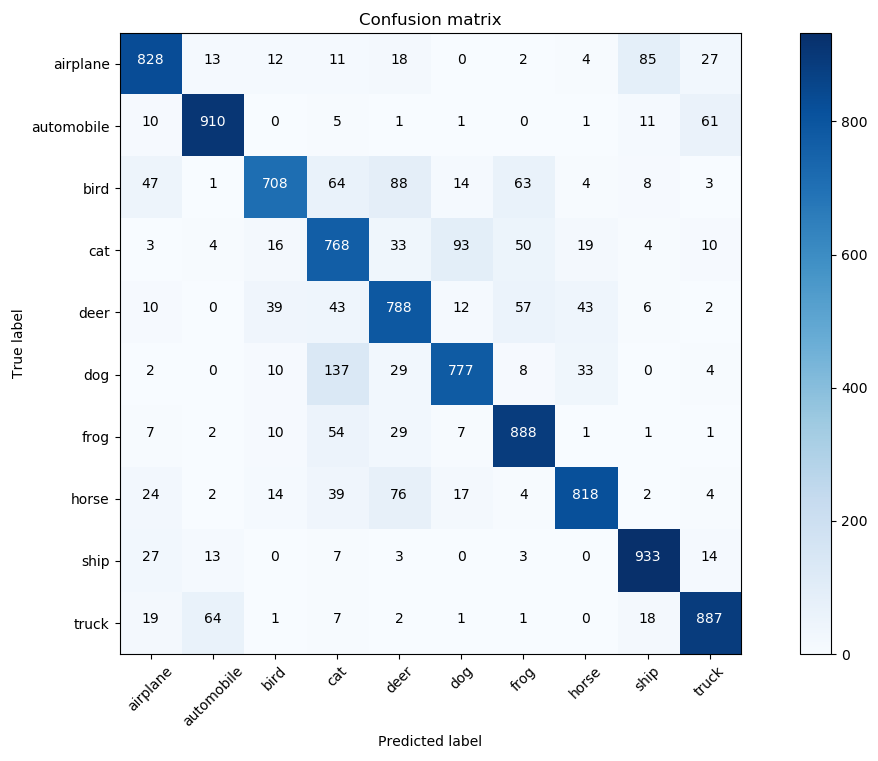}
  \caption{Confusion matrix on the CIFAR-10 dataset reported in \cite{hopfieldNetwork} We note that they report true labels on the y-axis of the confusion matrix.}
  \label{fig:hopfield_cm}
\end{figure}

\begin{figure}[!]
  \centering
  \includegraphics[width=3.2in]{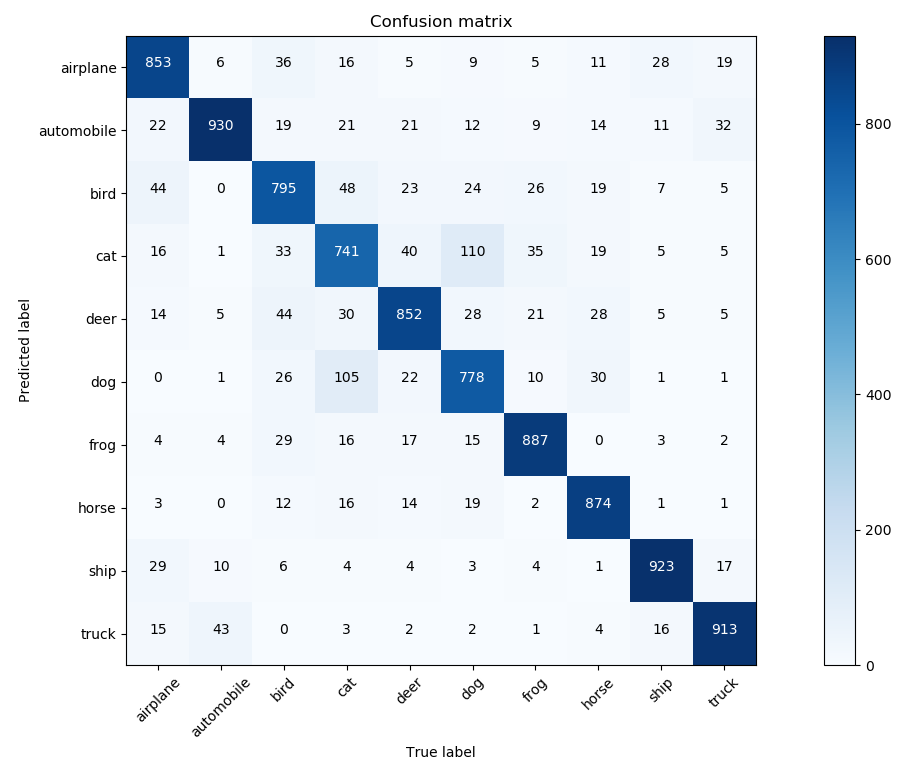}
  \caption{Confusion matrix on the CIFAR-10 dataset using our Res4Cif and GCML with parameter $\tau = 0.001$. Our method outperforms UL-Hopfield networks \cite{hopfieldNetwork} with significantly fewer parameters and without pre-training. We also note that in another experiment using $\tau = 0.009$, classification improved, as shown in Table \ref{tab:cifar_tau_acc}, implying further potential improvement.}
  \label{fig:gcml_cifar_cm}
\end{figure}

\begin{table}
  \renewcommand{\arraystretch}{1.3}
  \caption{CIFAR-10 Results for classification using GCML and different values of the $\tau$ hyperparameter. This threshold parameter is used to train the GCML structure as well as perform inference.}
\centering
\begin{tabular}{|c|c|c|}
\hline
Hyperparameter $\tau$ value & Percent correct & Training epochs\\
\hline
0.3 & 76.59 & 80\\
0.1 & 82.21 & 80\\
0.05 & 83.6 & 80\\
0.009 & \textbf{85.51} & 80\\
0.001 & 85.5 & 80\\
\hline
\end{tabular}
\label{tab:cifar_tau_acc}
  
\end{table}

Table \ref{tab:cifar_tau_acc} shows the performance of our method on CIFAR-10 at different values of the $\tau$ hyperparameter. Optimal values of this parameter depend on the the type of data normalization that is performed on the original data, and normalization methods of activated feature maps that are used to compute input to our attention function $Q$. In this experiment we did not perform any normalization on the input data and used simple \emph{min-max} normalization on activated feature maps. Figure \ref{fig:hopfield_cm} shows the confusion matrix on the CIFAR-10 dataset reported in \cite{hopfieldNetwork} and Figure \ref{fig:gcml_cifar_cm} shows our method's confusion matrix on the same data.

\subsection{COVID-19 Radiology Results}
The main goal of this work was to evaluate the potential benefit of attention mechanisms in identifying pulmonary conditions in chest radiographs that may be missed by spatial invariance of convolutional neural networks. Vision transformers are an active area of research and convolutional neural nets are currently the main approach for diagnostic assistance in the chest radiology domain \cite{butt2020deep,wang2020deep, zhou2018unet++,roy2020deep,hurt2020deep}, thus we focus on these techniques. Data in the radiology domain is not as well curated as in the natural imagery domain, with new diseases like COVID-19 presenting new classes to identify. This presents both, class imbalanced and data starved environments. Here we present our results of applying our method to the publicly available COVID-19 Radiology dataset \cite{covidDataKaggle1}.

For this experiment we randomly split the COVID-19 Radiology dataset into train, validation, and test partitions as shown in Table \ref{tab:covid_data}. The validation set is used to select the best set of weights during the model $G$ fine-tuning process, which is a standard practice in model training. Since validation data is never used to update weights of $G$ during training, in the second part of our experiment we utilize the validation set to further refine the GCML structure to investigate the effect of updating just the attention mechanism without fine-tuning the feature extractor model on this data. 

\begin{table}
  \caption{COVID-19 Dataset \cite{covidDataKaggle1} partitions. The validation partition is used as a model selection criteria during fine-tuning of the feature extractor model.}
  \renewcommand{\arraystretch}{1.3}
  \centering
  \begin{tabular}{|c|c|c|c|}
    \hline
    Data Class Samples & Training & Validation & Test\\
    \hline
    COVID-19 & 131 & 44 & 44\\
    No Finding & 804 & 269 & 268\\
    Viral Pneumonia & 807 & 269 & 268\\
    \hline
  \end{tabular}
  \label{tab:covid_data}
\end{table}

Since we have relatively few training points, we utilized transfer learning by fine-tuning a ResNet50 architecture that was pre-trained on the ImageNet dataset. The standard modification that is done during transfer learning is to remove the last connected layer that represents original classes and replace it with new target classes before fine-tuning with new data. One other modification that we performed was to reduce the final mapping resolution of the network from $7 \times 7$ to $5 \times 5$. This is done by adding a single convolutional layer with a kernel (filter) size of 3, padding of 0, and unit stride, while preserving the same number of in and out channels as the previous convolutional layer. This follows from simple ouput resolution calculus for a given dimension of a CNN convolutional layer, where for any equilateral input $i$, kernel size $k$, padding $p$, and stride $s=1$, the output resolution is given by $o = (i - k) + 2p + 1$.

We then fine-tuned all parameters of this architecture using the training partition of the COVID-19 dataset \cite{covidDataKaggle1} for 30 epochs, while using the validation partition to keep track of the best performing weights, selecting them as our final feature extractor model.  Fine-tuning was done using the following settings and hyperparameters: 
\begin{itemize}
  \item input was resized to $224 \times 224$ through a random resize crop and normalized using \emph{ImageNet} mean and standard deviation values
  \item \emph{Stochastic Gradient Descent} was used as our optimization function
  \item \emph{Cross Entropy Loss} was our training criterion
  \item \emph{learning rate} was set to $0.001$
  \item \emph{momentum} was set to $0.9$
\end{itemize}

For our next step we trained the GCML structure using the training partition of \cite{covidDataKaggle1} and our convolutional neural network from the previous step using several \emph{cam activation points} or $\tau$ values for 15 epochs each, with the best performing value on the test portion being $\tau = 0.05$. Training for multiple epochs was done because we used the same \emph{random resize crop} data transform that was done during training of our feature extractor model, presenting slight positional variations. Finally, we run our method on the test partition of the COVID-19 Radiology dataset keeping track of both classifications, standard CNN and GCML attention mechanism, as shown in paths $(a)$ and $(b)$ of Figure \ref{fig:gcml_flow}.

\begin{table}
  \caption{COVID-19 dataset \cite{covidDataKaggle1} combined results for accuracy, F1 score, and sensitivity using standard CNN, our GCML approach, and further tuned GCML$^T$ structure using the test partition of the COVID-19 Radiology dataset \cite{covidDataKaggle1}.} 
  \renewcommand{\arraystretch}{1.3}
  \centering
  \begin{tabular}{|c|c|c|c|}
    \hline
    METRIC (95\% CI) & CNN & GCML & GCML$^T$\\
    \hline
    Accuracy  & 0.948 $\pm$0.018 & 0.938 $\pm$0.020& 0.942 $\pm$0.019\\
   
    F1 Score & 0.973 $\pm$0.013& 0.968 $\pm$0.014& 0.970 $\pm$0.014\\
   
    Sensitivity & 0.962 $\pm$0.016& 0.955 $\pm$0.017& 0.958 $\pm$0.016\\
    \hline
  \end{tabular}
  \label{tab:rad_kaggle_results}
\end{table}



\begin{table}
  \caption{COVID-19 per-class classification accuracy (95\% CI) for standard CNN, our GCML approach, and further tuned GCML$^T$ structure using the test partition of the COVID-19 Radiology dataset \cite{covidDataKaggle1}. We also note that GCML$^T$ structure was further tuned on the validation partition, while our feature extractor used with GCML$^T$ was not further trained with the validation partition.}
  \renewcommand{\arraystretch}{1.3}
  \centering
  \begin{tabular}{|c|c|c|c|}
    \hline
    METHOD &COVID-19 & No Finding & Viral Pneumonia\\
    \hline
    CNN & 1.0 & 0.978 $\pm$0.018& 0.911 $\pm$0.034\\
   
    GCML & 1.0 & 0.929 $\pm$0.031 & 0.937 $\pm$0.029\\
    
    GCML$^T$ & 1.0 & 0.937 $\pm$0.029 & 0.937 $\pm$0.029\\
    \hline
  \end{tabular}
  \label{tab:covid_kaggle_class_acc}
\end{table}

During the training phase of the GCML structure, we explicitly made the normalization step, shown on line 7 of Algorithm \ref{alg:train}, optional. This allows us to further train a given GCML structure on new data without having to track distribution statistics of the original training data. For our next experiment we evaluated the effect of further training the GCML structure on additional data points without further tuning the convolutional neural network used as the feature extractor. We used the validation portion of \cite{covidDataKaggle1} to further train only the GCML structure with the same threshold $\tau = 0.05$. We used 44 samples of COVID-19, 269 samples of No Findings, and 269 samples of Viral Pneumonia images to train for 5 epochs. As we described earlier, forgoing the optinal normalization step in Algorithm \ref{alg:train} allows for this functionality. We then ran the test partition, which neither the convolutional neural network nor the GCML structure have been trained on, to obtain results for combined accuracy, F1 score, and sensitivity.  Table \ref{tab:rad_kaggle_results} shows our results using three models, \emph{CNN} or standard classification using our convolutional neural network, \emph{GCML} using our attention mechanism, and finally GCML$^T$, where we further trained the attention mechanism using the validation partition.

The combined results in Table \ref{tab:rad_kaggle_results} show several encouraging results.  First, the attention mechanism performed similarly well with a combined accuracy of 0.938 as compared to the traditional classification method of convolutional neural networks with a combined accuracy of 0.948.  Second, by further tuning the GCML structure, to obtain GCML$^T$, with a small number of data points we were able to improve all metrics, bringing combined accuracy to 0.942.  This is encouraging because tuning this structure is significantly faster than tuning a convolutional neural network. Finally, once we drilled down to class level accuracy, we observe that GCML improves the standard CNN classification on the Viral Pneumonia class by about 3 percent, as shown in Table \ref{tab:covid_kaggle_class_acc}. This was significant for two reasons, first \emph{it empirically verified our main hypothesis that our auxiliary attention mechanism could identify cases of pulmonary conditions that the standard CNN classification missed}. Second, these results support our intuition of the additive nature of standard CNN classification by showing that this method outperformed the attention method on the \emph{No Findings} class, as spatial relations between activated features matter significantly less, as expected. 

\begin{figure*}[t!]
  \centering
  \begin{subfigure}[t]{0.333\textwidth}
    \centering
    \includegraphics[height=2.3in]{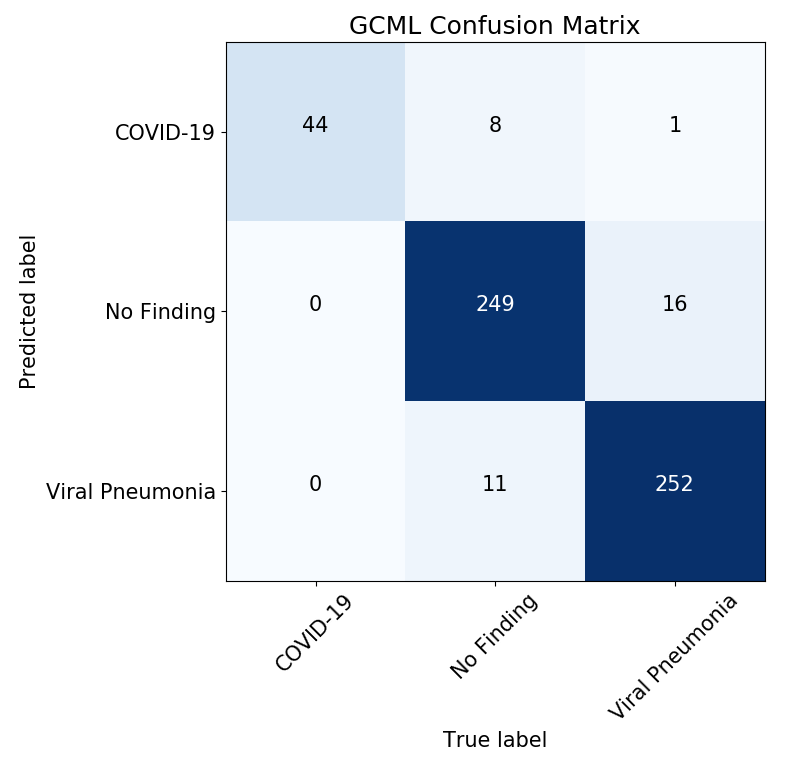}
    \caption{}
  \end{subfigure}%
  \begin{subfigure}[t]{0.333\textwidth}
    \centering
    \includegraphics[height=2.3in]{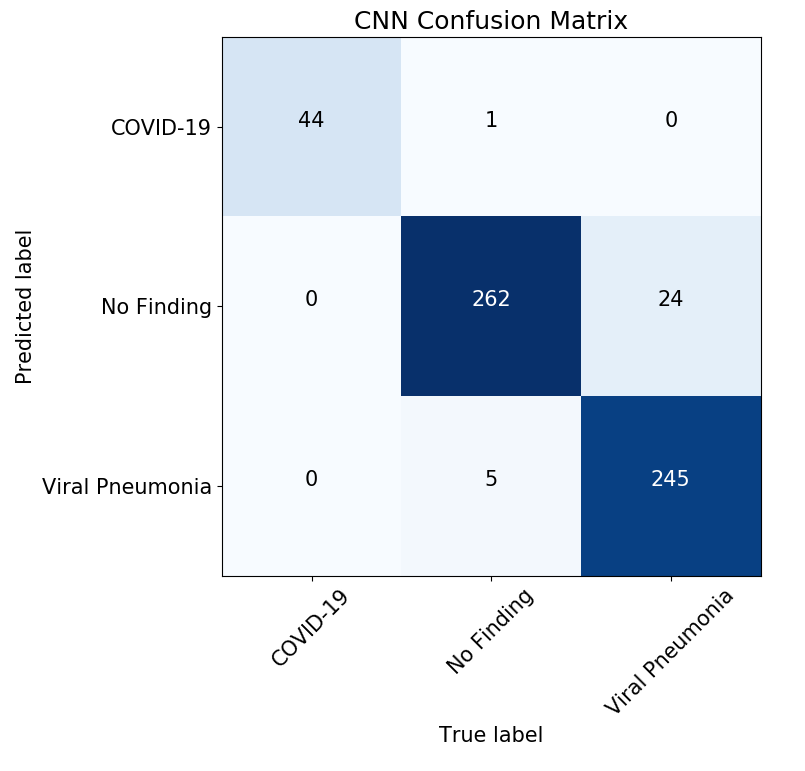}
    \caption{}
  \end{subfigure}%
  \begin{subfigure}[t]{0.333\textwidth}
    \centering
    \includegraphics[height=2.3in]{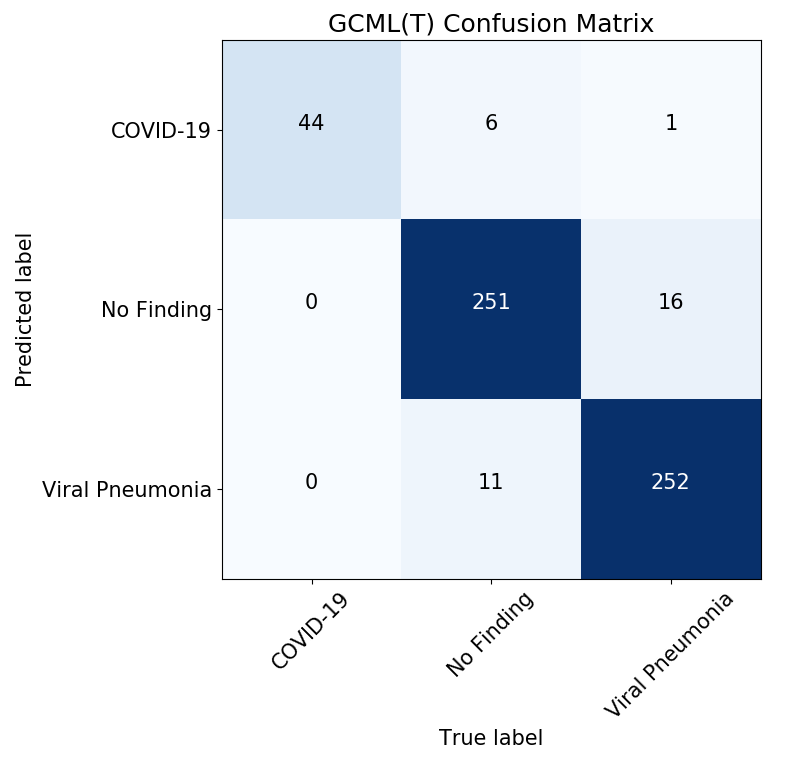}
    \caption{}
  \end{subfigure}
  \caption{Confusion matrices for evaluation results using the test partition of the COVID-19 Radiology dataset \cite{covidDataKaggle1}. The three results represent the following: (a) classification using our GCML attention mechanism, (b) classification using the final connected layer of our fine-tuned Convolutional Neural Network, and (c) classification using our GCML$^T$ attention mechanism further trained using the validation partition of \cite{covidDataKaggle1}. Our first observation is that our attention mechanism (a) is able to identify pulmonary conditions that the standard CNN (b) missed. This provides empirical evidence to our hypothesis for the utility of attention mechanisms in the chest radiology domain. Second, by further training our GCML structure using the validation partition, which was only used as a model selection criteria during the training of our feature extractor, performance was further improved (c). This was done without further training of the CNN feature extractor. These  results suggest that hybrid CNN plus attention-based ensemble techniques, that utilize different inductive biases, provide a promising set of approaches to incorporating global interactions between dispersed features of pulmonary diseases that are manifested in chest radiographs.}
  \label{fig:xray_results_confusion}
\end{figure*}

Due to the absense of object level labels, or bounding boxes labeling localized manifestations of pulmonary conditions in training data, it is difficult to identify feature specific intersections or lack thereof between images that CNN and GCML classified correctly or incorrectly. But even with only image level labels for training, where we know that the condition is present but have no localized information of its manifistation, the confusion matrices in Fig \ref{fig:xray_results_confusion} clearly show GCML's improved performance on the \emph{Viral Pneumonia} class, while both methods performed the same on the \emph{COVID-19} class. 

\subsection{Radiology Generalization Performance}
For our final experiment we examined our method's ability to generalize to a separate dataset \cite{cohen2020covid} containing X-ray samples of patients diagnosed with COVID-19, No Findings, and Pneumonia, both viral and bacterial. We created this dataset by extracting images from a publicly available repository \cite{cohen2020covid} in order to create a test partition that was comparable in size to the train partition we used in training our method. The main motivation for this experiment was an observation that many instances of work related to diagnosing COVID-19 and other pulmonary conditions using chest radiographs omit external generalization experiments, as reported in Roberts et al. \cite{Roberts2021Nature}.

Table \ref{tab:generalization_covid_data} shows the generalization test partition that we created using \cite{cohen2020covid}. Neither our feature extractor nor the GCML structure were trained using this data. We then tested this data using GCML classification, using $\tau = 0.05$. The only transformations that we performed during inference were resizing the image to $224 \times 224$, and normalization of the tensorized image to \emph{ImageNet} values for mean and standard deviation.

\begin{table}
  \caption{Generalization test dataset for COVID-19 Radiology Data derived from \cite{cohen2020covid}. Neither our feature extractor $G$, nor the GCML structure $S$ were trained on this data.}
  \renewcommand{\arraystretch}{1.3}
  \centering
  \begin{tabular}{|c|c|c|}
    \hline
    COVID-19 & No Finding & Pneumonia (Viral and Bacterial)\\
    \hline
    133 & 949 & 390\\
    \hline
  \end{tabular}
  \label{tab:generalization_covid_data}
\end{table}

\begin{table}
  \caption{Generalization dataset \cite{cohen2020covid}  results at 95\% confidence interval. Combined accuracy, F1 score, and sensitivity are shown using our GCML classification approach. Last row shows per-class accuracy for the same classifier.}
  \renewcommand{\arraystretch}{1.3}
  \centering
  \begin{tabular}{|c|c|c|c|}
    \hline
     (95\% CI) & GCML &   &  \\
    \hline
    Accuracy  & 0.940 $\pm$0.012 &  & \\
   
    F1 Score & 0.969 $\pm$0.009 & & \\
    
    Sensitivity & 0.955 $\pm$0.011 &  &  \\
    \hline
      & COVID-19 & No Finding & Pneumonia \\
    \hline
    Class Accuracy & 0.985 $\pm$0.021 & 0.9294 $\pm$0.016 & 0.951 $\pm$0.021\\
    \hline
  \end{tabular}
  \label{tab:rad_generalization_results}
\end{table}

\begin{figure}
  \centering
  \includegraphics[width=3.2in]{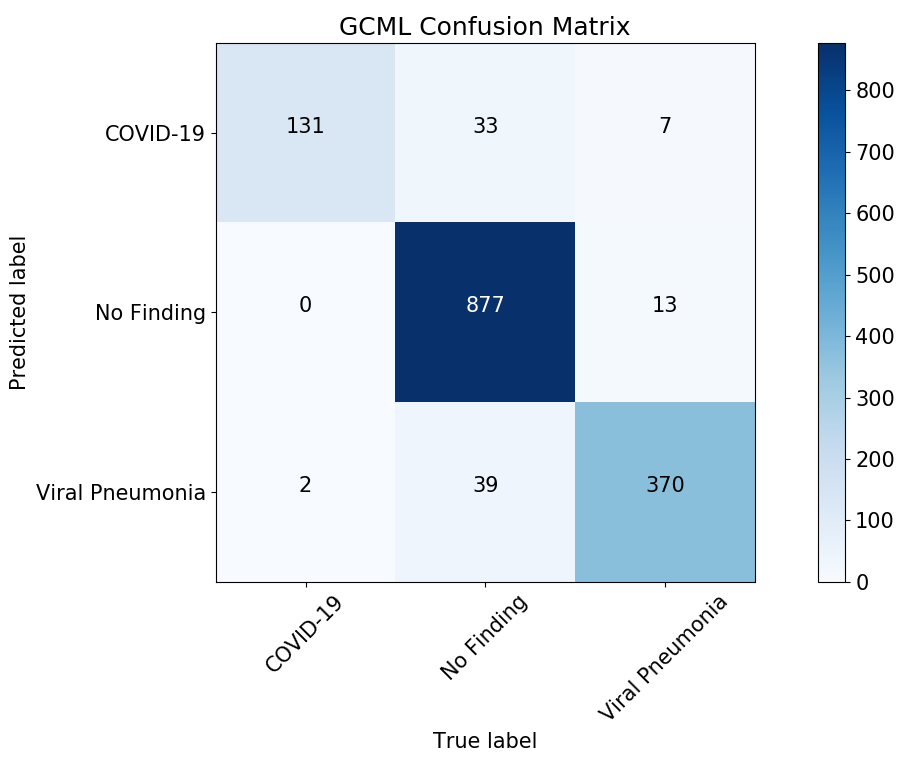}
  \caption{Confusion matrix for predictions made on the radiology generalization test dataset, created from \cite{cohen2020covid}, using the GCML auxiliary attention mechanism. As noted, no training was performed using this dataset of neither the feature extractor nor the GCML attention mechanism.}
  \label{fig:generalize_gcml_cm}
\end{figure}

Table \ref{tab:rad_generalization_results} shows our results, including per class accuracy, while Fig \ref{fig:generalize_gcml_cm} shows the confusion matrix. Similar to our initial results on \cite{covidDataKaggle1}, we saw our attention mechanism perform well at identifying pulmonary conditions, both COVID-19 and Pneumonia, and slightly worse on the No Findings class. Further examining per-class performance we see that sensitivy rates for both COVID-19 and Pneumonia are high, making the approach effective at detecting infected patients, represented in data never seen by the model during training. These generalization results also increase our confidence in that pertinent features of pulmonary conditions are being discovered by the model, as opposed to spurious artifacts such as X-ray markings related to the origin or medium of radiographs.

\subsection{Confidence Level on Hypothesis Test}
To evaluate our main hypothesis regarding our attention mechanism outperforming the \emph{standard} CNN when classifying pulmonary conditions, we perform a hypothesis test \cite{Johnson_stat} on the two classifers as they pertain to pneumonia classification.  Both classifiers performed equally well on COVID-19 classification. Let $p_1 = 0.911$ be CNN accuracy and $p_2 = 0.937$ be GCML accuracy shown in Table \ref{tab:covid_kaggle_class_acc}. Let $n = 268$ be the number of samples of viral pneumaonia images in the test partition of \cite{covidDataKaggle1}. Then, we calculate the test statistic $Z$ as:

\begin{equation}
  Z = \frac{p_1 - p_2}{\sqrt{2\hat{p}(1-\hat{p})/n}}
\end{equation}

\noindent where $\hat{p} = (245 + 252)/2(268)$, with 245 and 252 being numbers of correct classifications from both classifiers, as shown in the confusion matrices in Figure \ref{fig:xray_results_confusion}.  To show that $p_1 < p_2$, or that $p_2$ is better than $p_1$, we need to show $Z < -z_{\alpha}$, where $z_{\alpha}$ is obtained from a standard normal distribution pertaining to a significance level $\alpha$. Given our sample size, we compute $Z = -1.1607$. Our test statistic is slightly better than the $Z$ value of 1.15035 for a 75\% confidence interval. Thus, we can state with 75\% confidence level that GCML is more accurate than standard CNN for classifying pneumonia cases.

\section{Conclusion}
Detection of pulmonary disease using Artificial Intelligence techniques is an emergent field \cite{ALBAHRI20201381}, driven not just by academic curiosity, but a real need to improve accessibility and speed of diagnosis. Well studied approaches to image classification need to be analyzed and improved to be effective in real world applications, where data can be scarce, noisy, and novel. Emergence of new diseases, such as COVID-19, exposed weaknesses in such applications of machine learning methods, but also highlighted potential benefits and promise to society if they were to be successfuly addressed \cite{Roberts2021Nature}. 

In this work, we developed a method to improve classification performance of pulmonary diseases in chest X-rays by expanding the perceptive awareness of convolutional neural networks via GCML, our new attention mechanism. We showed that our auxiliary attention mechanism improved sensitivity of pulmonary disease classifers by accounting for spatial interrelations of features globally. Our initial results show a promising direction of research towards improving existing methods of image classification for diagnostic assistance of pulmonary diseases.

There are two main limitations of our initial approach.  First, the current GCML datastore learns a conditional discrete probability distribution of a given class, more specifically the use of the $\tau$ parameter to threshold activation values of the input to attention function $Q$, results in a binomial distribution. This results in some information loss compared to a continuous distribution.  Second, as described in Section \ref{sec:approach}, we must reduce the final mapping resolution of our CNNs to manage the size of learned distributions, which essentially increases the area of individual image patches for which we learn global correlations. This can have a smoothing effect on multiple disease manifestations detected in a single region, causing it to be treated as a single manifistation. To address these limitations, we plan on improving the attention mechanism by utilizing normalized continuous values of class activation maps without a threshold parameter by learning a continuous multivariate distribution, such as a Dirichlet distribution. This will allow for image patches with smaller areas, resulting in more patches, to be used in learning global relations among even smaller manifestations.

Additionally, we can improve positional probability distributions learned by GCML structure by reducing class noise due to class overlapping data in the chest X-ray domain. Similar classes result in noisy class activation maps that contain many overlapping activations \cite{verenich2020improvingICMLA}.  By better extracting class relevant activations, we can learn tighter conditional probability distributions for appropriate classes of diseases.

We believe that improving well established techniques for image classification towards domain-specific applications, such as pulmonary disease classification, is well worth the effort. By utilizing a strong technical foundation of CNNs, progress towards a useful diagnostic aid can be accelerated, and outcomes of patients, especially in regions where access to sophisticated laboratory diagnostics is limited, can be improved.  To that end, we make the reference implementation of our attention mechanism freely available, including source code and models \cite{gcmlArtifacts}.


%




\ifCLASSOPTIONcaptionsoff
  \newpage
\fi



\bibliographystyle{IEEEtran}
\bibliography{gcml}
%



\end{document}